\documentclass[runningheads]{llncs}
\usepackage{graphicx}
\usepackage{tikz}
\usepackage{comment}
\usepackage{amsmath,amssymb} % define this before the line numbering.
\usepackage{color}

\usepackage[accsupp]{axessibility}  % Improves PDF readability for those with disabilities.
\usepackage{amsmath}
\usepackage{amssymb}
\usepackage{booktabs}
\usepackage{multicol}
\usepackage{multirow}
\usepackage{makecell}
\usepackage{algorithm}
\usepackage{algpseudocode}
\usepackage{bbding}
\usepackage{hyperref}
\usepackage{subfigure}
\usepackage{xcolor,colortbl}
\usepackage{colortbl}
\definecolor{Gray}{gray}{0.85}
\definecolor{LightCyan}{rgb}{0.88,1,1}
\definecolor{emerald}{rgb}{0.0, 0.8, 0.5}
\definecolor{harvestgold}{rgb}{1, 0.77, 0.05}
\begin{document}
% \linenumbers
\pagestyle{headings}
\mainmatter
\def\ECCVSubNumber{12}  % Insert your submission number here

\title{TransFiner: A Full-Scale Refinement Approach for Multiple Object Tracking} % Replace with your title

\begin{comment}
\titlerunning{ECCV-22 submission ID \ECCVSubNumber} 
\authorrunning{ECCV-22 submission ID \ECCVSubNumber} 
\author{Anonymous ECCV submission}
\institute{Paper ID \ECCVSubNumber}
\end{comment}
%******************

%begin{comment}
\author{
	Bin Sun
}
%
%\authorrunning{H. qiu, Y. Ma, Z. Li, S. Liu and J. Sun}
\institute{
	China University of Geosciences
}
%\end{comment}
%******************
\maketitle

\begin{comment}
	tips for writing abstract:
		State the issue to be discussed
		Give a brief background to the issue
		Brief description of what you are doing about it
		Implications/outcomes: why is what you've done important?
\end{comment}
\begin{abstract}
	Multiple object tracking (MOT) is the task containing detection and association. Plenty of trackers have achieved competitive performance. Unfortunately, for the lack of informative exchange on these subtasks, they are often biased toward one of the two and underperform in complex scenarios, such as the inevitable misses and mistaken trajectories of targets when tracking individuals within a crowd. This paper proposes TransFiner, a transformer-based approach to \textit{post-refining} MOT. It is a generic attachment framework that depends on \textit{query pairs}, the bridge between an \textit{original tracker} and TransFiner. Each query pair, through the \textit{fusion decoder}, produces refined detection and motion clues for a specific object. Before that, they are feature-aligned and group-labeled under the guidance of tracking results (locations and class predictions) from the original tracker, finishing tracking refinement with focus and comprehensively. Experiments show that our design is effective, on the MOT17 benchmark, we elevate the CenterTrack from $ 67.8\% $ MOTA and $ 64.7\% $ IDF1 to $ 71.5\% $ MOTA and $ 66.8\% $ IDF1.
\end{abstract}

\section{Introduction}
\label{sec:Introduction}

\begin{comment}
	{sun2020transtrack}
	{meinhardt2021trackformer}
	{chen2022patchtrack}
	{zhu2021looking}
	{cai2022memot}
	{xu2021transcenter}
	
	memory mechanism will 1) bring the difficulty to remove the false postive once produced 
	2) 
\end{comment}

\begin{comment}
	\begin{figure}[t]
		\begin{center}
			\includegraphics[width=0.9\linewidth]{figure/intro_transfiner.png}
		\end{center}
		\caption{Inspiration of TransFiner. Pipelines of preparing queries for the decoder. (a) Track queries from previous frame, directly \cite{meinhardt2021trackformer,sun2020transtrack} or enhanced by features from current frame \cite{chen2022patchtrack}. (b) History buffer is responsible for producing track queries \cite{zhu2021looking,cai2022memot}. (c) Ours. As a post-refinement framework, we directly enhance the object queries with aligned image features obtained under the guidance of the original tracker.}
		\label{fig:intro_difference}
	\end{figure}
\end{comment}
\begin{figure}[t]
	%\centering
	\subfigure[Track queries from the previous]{
		\begin{minipage}[b]{0.25\textwidth}
			\includegraphics[width=0.8\textwidth]{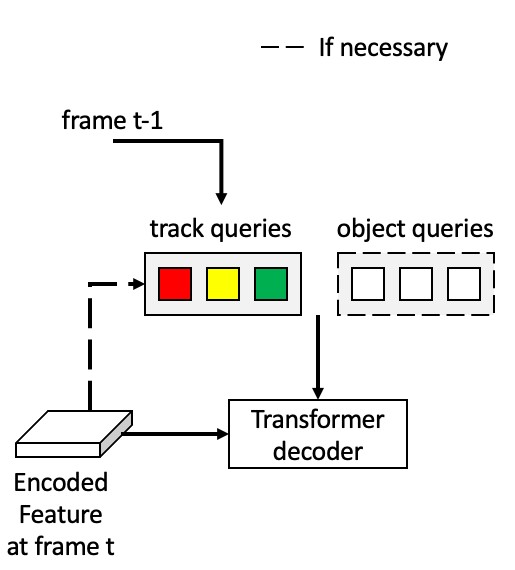}
		\end{minipage}
		\label{fig:intro_difference:fromprevious}
	}
	\subfigure[Track queries from the history buffer]{
		\begin{minipage}[b]{0.25\textwidth}
			\includegraphics[width=0.8\textwidth]{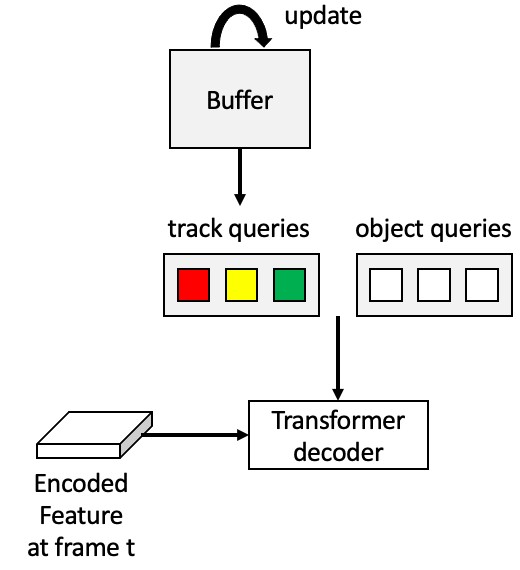}
		\end{minipage}
		\label{fig:intro_difference:fromhistorybuffer}
	}
	\subfigure[Object query pairs enhanced by the original tracker (Ours)]{
		\begin{minipage}[b]{0.4\textwidth}
			\includegraphics[width=0.8\textwidth]{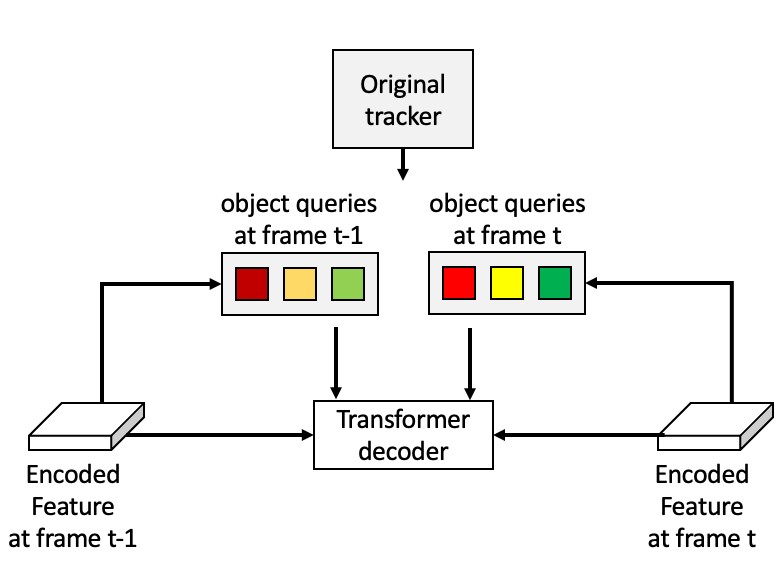}
		\end{minipage}
		\label{fig:intro_difference:ours}
	}
	\caption{\textbf{Pipelines of preparing queries for the decoder.} \ref{fig:intro_difference:fromprevious} Track queries from the previous frame, directly \cite{meinhardt2021trackformer,sun2020transtrack} or enhanced by features from the current frame \cite{chen2022patchtrack}. \ref{fig:intro_difference:fromhistorybuffer} History buffer is responsible for producing track queries \cite{zhu2021looking,cai2022memot}. \ref{fig:intro_difference:ours} Ours. As a post-refinement framework, we fill the object query pairs across frames with encoded features obtained under customized guidance from the original tracker.}
	\label{fig:intro_difference}
\end{figure}

Multiple object tracking (MOT) refers to linking identical detections across frames and primarily exists in the form of  two mainstream paradigms, namely tracking-by-detection (TBD) and joint detection and tracking (JDT). TBD approaches \cite{bewley2016simple,vatralonline,leal2016learning,fang2018recurrent,pang2021quasi} split the MOT into two separate stages, including detection and association. JDT, alternatively, solves the MOT problem in unified ways via constructing a tracking-related structure \cite{feichtenhofer2017detect,zhou2020tracking,shuai2021siammot,wu2021track} within or adjusting the output objective of the particular branch \cite{bergmann2019tracking} of the existing detectors. From an additionally emerging paradigm, transformer-based MOT formulations \cite{sun2020transtrack,meinhardt2021trackformer,chen2022patchtrack,zhu2021looking,cai2022memot,xu2021transcenter} also finish tracking satisfactorily. Nevertheless, these methods still struggle with intricate scenarios, such as several objects passing each other and patches of crowded objects, which lead to either high false alarms (or high miss rate) and degrade association simultaneously. On the other hand, with DETR \cite{carion2020end}, end-to-end object detection is realized through object queries and Hungarian loss, facilitating individual-separate detection (e.g., without the need for NMS).

% As a result, achieving competitiveness both in detection and association is challenging amid MOT.  start to describle double decoder with fusion that can one to one association and detecction. , embedding the thought of DETR-like methods \cite{carion2020end,li2022dn,zhang2022accelerating,zhu2020deformable,yao2021efficient,meng2021conditional}, As shown in Figure \ref{fig:intro_difference:ours}, instead of track queries from the previous \ref{fig:intro_difference:fromprevious} or buffer \ref{fig:intro_difference:fromhistorybuffer} locations are utilized to extract encoded features to freshly initialize object query pairs for detection (frame t) and association (frame t-1)

In light of these, we show how to build a generic and targeted framework for refining MOT, referred to as TransFiner, a transformer-based refinement approach. Unlike most related work, DRT \cite{wang2021drt} refines MOT patch by patch, which indeed improves detection but hardly promotes association (even degrades it according to the IDF1 reported in experiments \cite{wang2021drt}). We, instead, take a full-scale approach by enriching query pairs guided by the original tracker (Figure \ref{fig:intro_difference:ours}), refinement then is a fine-tuning process for query pairs without scope restriction.

As summarized in Figure 1, the existing transformer-based MOT formulations \cite{meinhardt2021trackformer,sun2020transtrack,chen2022patchtrack,zhu2021looking,cai2022memot,xu2021transcenter} primarily accomplish tracking via the tracklet record (e.g., track query). Instead, we use freshly initialized query pairs (i.e., separately for detection and association) for every shot. With this design, we note that a competitive tracking refinement can be achieved while less affected by the formerly poor tracking predictions.

TransFiner takes originally estimated object locations, class predictions, and two successive frames as inputs, predicting detections (frame $ t $) and association clues containing motions of center and box (mapping detections from frame $ t $ to frame $ t-1 $). These are achieved via TransFiner's \textit{query pairs} plus \textit{fusion decoder}. The latter consists of the fusion attention module and dual-decoder. Specifically, fusion attention is responsible for the interaction between query pairs, while dual-decoder is assigned to take care of these two separately.

In order to better utilize information from the original tracker, predictions are categorized into qualified and poor ones in terms of their class scores. Together with learnable label embeddings, TransFiner finishes targeted refinement with different focuses of query embeddings on various estimations in parallel. During training, we additionally refer to ground-truth objects when pre-assigning refinement targets to original estimations with \textit{close} distance, avoiding instability introduced in layer-wise Hungarian matching when refining.
% roi align with tgt

% TODO: further determine when getting test results
Experiments show that tracker refined by TransFiner are robust enough to revisit compelling performance. With TransFiner's refinement, CenterTrack achieves 71.5\% MOTA, 66.8\% IDF1 on the MOT17 benchmark.

\section{Related work}
\label{sec:relatedwork}

\subsection{Association in tracking}
Motion and appearance are two crucial references when linking detections between frames. Several works rely solely on motions, guiding objects to the next frame \cite{bewley2016simple,vatralonline,bergmann2019tracking,shuai2021siammot,feichtenhofer2017detect} or moving them backward \cite{zhou2020tracking,xu2021transcenter} to search for associated ones. Some \cite{lu2020retinatrack,pang2021quasi} take advantage of appearance features to match interframe objects by computing similarity scores between feature embeddings. Naturally, combining both in association \cite{leal2016learning,wojke2017simple,son2017multi,fang2018recurrent,zhang2020fairmot,wu2021track,du2022strongsort} is also widely explored.

Another recent popular trend builds on transformer \cite{vaswani2017attention}, packaging the preceding information into high-level embeddings (e.g., track queries \cite{meinhardt2021trackformer,sun2020transtrack,chen2022patchtrack,zhu2021looking,cai2022memot}). These embeddings are then processed together with the current information \cite{meinhardt2021trackformer,zhu2021looking,chen2022patchtrack,cai2022memot}, or they serve as the initialization in the latest detection \cite{sun2020transtrack}, handling association problems via \textit{another detection shot}. Our method extends this trend by injecting freshly aligned and grouped encoded features to query pairs focused on joint prediction (refinement) of detections and corresponding motions for association, which is completed in one run. Furthermore, we package information from centers and boxes into motions, facilitating precise association even among crowds.

\subsection{DETR and its variants}
DETR \cite{carion2020end} handles object detection in an end-to-end manner. This primarily benefits from the transformer’s attention mechanism and the introduction of object query, unfortunately, two dominating factors contribute to the slow convergence of DETR (e.g., It takes 500 training epochs to achieve a competitive performance). To be specific, several variants  \cite{zhu2020deformable,gao2021fast,meng2021conditional} improve the attention module by designing mechanisms to constrain the interaction fields (e.g., sampling points \cite{zhu2020deformable}, additional spatial attention weight \cite{gao2021fast,meng2021conditional}), easing the match burden in comparison to the inefficient global search from DETR. Differently, object query alignment is studied in  \cite{yao2021efficient,zhang2022accelerating}, with the retrieval of the queries from encoded features showing effectiveness in accelerating convergence.

We build upon Deformable DETR \cite{zhu2020deformable}. Specifically for tracking (refinement), we construct a fusion decoder composed of fusion attention and a dual-decoder. Two decoders are connected through the fusion attention module, an additionally masked self-attention mechanism, ensuring effective intercommunication of query pairs. It is noteworthy that query pairs are iteratively aligned based on the inherent variable \textit{reference locations}. Repetitive refinement is then realized through consecutive updates of the pairs via decoder layers.

\subsection{Refinement}
By exploring the joint space of inputs and outputs, refinement can generally be divided into multi- and single-step approaches. The former involves iterative correction \cite{carreira2016human,tang2021recurrent} and cascaded rectification \cite{gidaris2017detect}. Contrastively, the latter simply attaches an independent module to the original model \cite{chen2018cascaded,fieraru2018learning,tang2021look,moon2019posefix,wang2021drt,yang2021remot}, yielding the refined results in a single pass.

MOT refinement focuses on optimizing detections and associations, and existing methods \cite{yang2021remot,wang2021drt} fall into the second category outlined above. ReMoT \cite{yang2021remot} enhances the tracklets of objects through a \textit{split-then-merge} strategy, reducing the identity switches, which, however, are not primary causes of performance degradation. Alternatively, DRT \cite{wang2021drt} refines the detection results from ambiguous patches, resulting in decent improvements. Nevertheless, due to the patch-based nature, the scope of post-processing is limited to a predefined area, making it different from full-scale refinement and failing to strengthen association performance effortlessly, over the original tracker. These inspire the design of TransFiner, a full-scale and single-step approach to refine MOT on detection and association.

\section{Preliminaries}
\label{sec:Preliminaries}

\noindent \textbf{Original tracker.} Generally taking a subset from frames $ \left\{I^t\right\}^{T}_{t=1} $ (up to the last frame $ I^T $ of video sequence) as inputs, \textit{original tracker} refers to the tracker whose predictions are to be refined. Outputs from the origin are 
$ \left\{ \hat{o}^{t}_{i} \right\}_{i=0}^{K-1} $, where $ K $ predictions are extracted from the post-processing with \textit{laxer} output settings (e.g., lower objectness score threshold). Let $ \hat{o}^{t}_{i} = (\hat{c}^{t}_{i}, \hat{b}^{t}_{i}, \hat{a}^{t}_{i}) $, $ \hat{c}^{t}_{i} $ and 
$ \hat{b}^{t}_{i} $ respectively indicate the classification score, as well as bounding box of object $ i $ out of the $ K $ predictions in frame $ t $. 
$ \hat{a}^{t} $ represents the association clues (e.g., motions \cite{bewley2016simple,feichtenhofer2017detect,bergmann2019tracking,vatralonline,shuai2021siammot} or feature embeddings \cite{lu2020retinatrack,zhang2020fairmot,pang2021quasi}) linking objects across frames.

\noindent \textbf{Refinement.} Let encoded image features from frame $ t $ and $ t-1 $ be the $ F^t $ and $ F^{t-1} $, respectively. Performing refinement on 
$ \left\{ \hat{o}^{t}_{i} \right\}_{i=0}^{K-1} $ contributes to 
$ \left\{ \tilde{y}^{t}_{i} \right\}_{i=0}^{N-1} $, where 
$ \tilde{y}^{t}_{i} = (\tilde{c}^{t}_{i}, \tilde{b}^{t}_{i}, \tilde{a}^{t}_{i}) $, and $ N $ is the number of queries in decoder. $ \tilde{a}^{t}_{i} $ denotes the motion of object $ i $ between frames. Additionally, TransFiner is built upon Deformable DETR \cite{zhu2020deformable}, whose decoder relies on the initial reference locations $ init\_ref $ to make final predictions.

\begin{figure}[t]
	\begin{center}
		\includegraphics[width=0.65\linewidth]{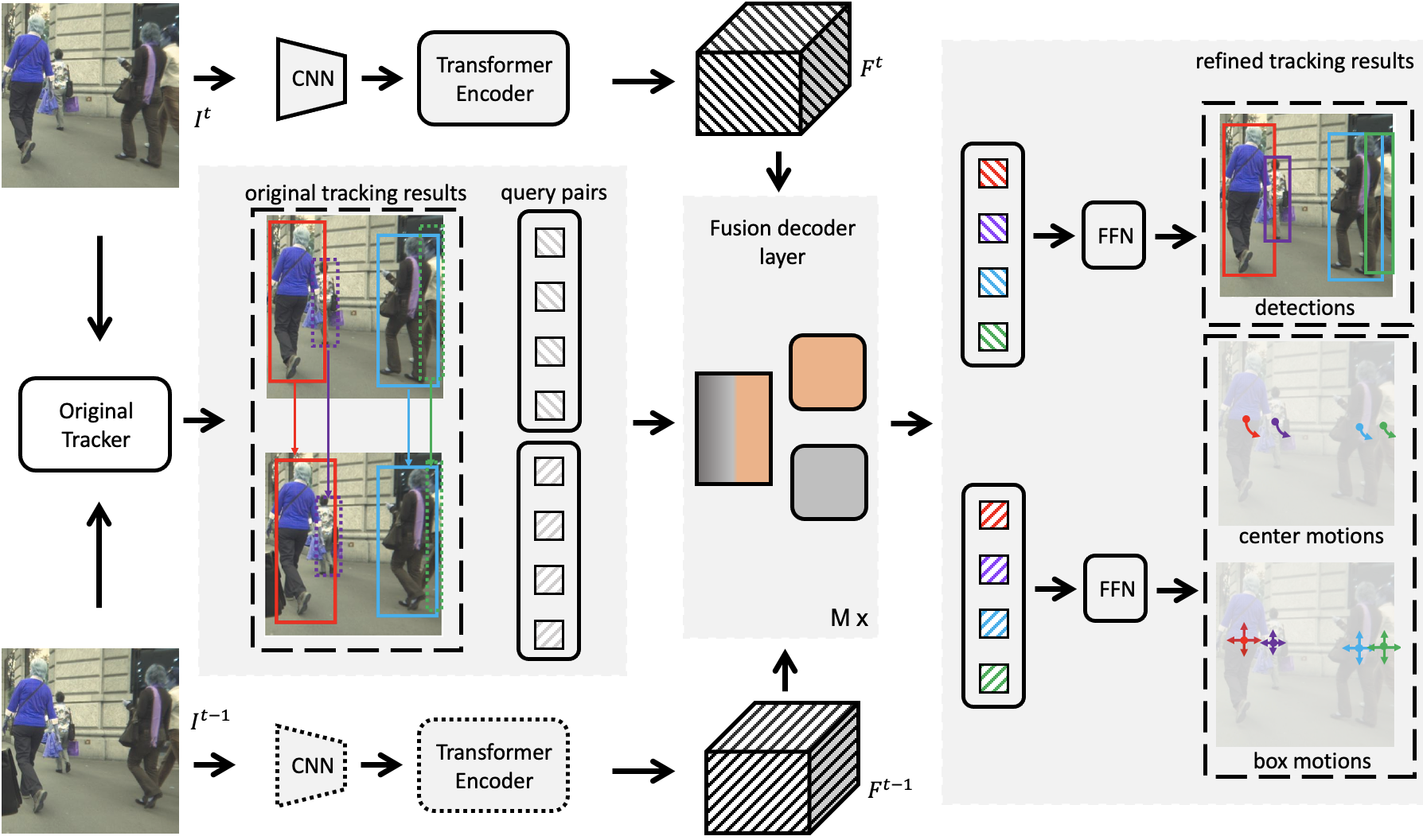}
	\end{center}
	\setlength{\belowcaptionskip}{-14pt} 
	\caption{\textbf{Refining tracker via TransFiner.} Encoded features $ F^t $ and $ F^{t-1} $ produced by the CNN backbone and encoder, original tracking results, and plain query pairs serve as inputs for the fusion decoder (i.e., $M$ fusion decoder layers). Following the FFN module, query pairs $ (Q^t, Q^{asso}) $ from fusion decoder transform into detections and motions. For the original results, boxes ignored by post-processing are in dotted form and are partially picked for illustration. The dotted CNN and the Encoder indicate that weights are shared with the solid ones.}
	\label{fig:overall_transfiner}
\end{figure}

\section{MOT refinement driven by transformer}
\label{sec:method}

\subsection{Why transformer in refinement}
Based on DETR \cite{carion2020end} and its derivations \cite{li2022dn,zhang2022accelerating,zhu2020deformable,yao2021efficient,meng2021conditional}, we show transformer's superiorities over convolutional neural networks in post-refinement in the following ways: (1) In DETR-like methods, stacked decoder layers gradually rectify predictions, resembling the refinement process; (2) The object query is regarded as the \textit{complex} of the corresponding target, the initialization of which, under the guidance of initial predictions, is finished with fetching specific image features, enabling targeted refinement; (3) Inspired by training with joint denoising and matching \cite{li2022dn}, refinement with transformer can be cast into two parallel processes: denoising qualified predictions and rematching for the poor ones. In the following sections, we describe how TransFiner incorporates these characteristics.

\subsection{Framework of TransFiner upon query pair}
\label{sec:method:framework}
A core concept in TransFiner is query pairs $ (Q^t, Q^{asso}) $. $ Q^t $ detects, and $ Q^{asso} $ produces \textit{related} motions (\textit{related} means each pair should take a specific object). Shown in Figure \ref{fig:overall_transfiner}, query pairs propagate within the fusion decoder. Thus, framework of TransFiner can be divided into three parts: decoder's inputs, decoding, and decoder's predictions over query pairs. For inputs, we package encoded features $ F^t $ and $ F^{t-1} $, original results $ \hat{o}^t $, and plain query pairs. Decoding, after targeted initialization on query pairs under $ \hat{o}^t $, focuses on fusing pairs and separately processing queries for detection ($Q^t$) and association ($Q^{asso}$), separately contributing to target estimations of frame $ t $ (i.e., $  (\tilde{b}^t, \tilde{c}^t) $) and association clues $ \tilde{a}^t $ as target motions of center and box relative to that of the previous frame.

\subsection{TransFiner's fusion decoder layer}
\label{sec:method:decoderlayer}

\begin{figure}[t]
	\begin{center}
		\includegraphics[width=0.85\linewidth]{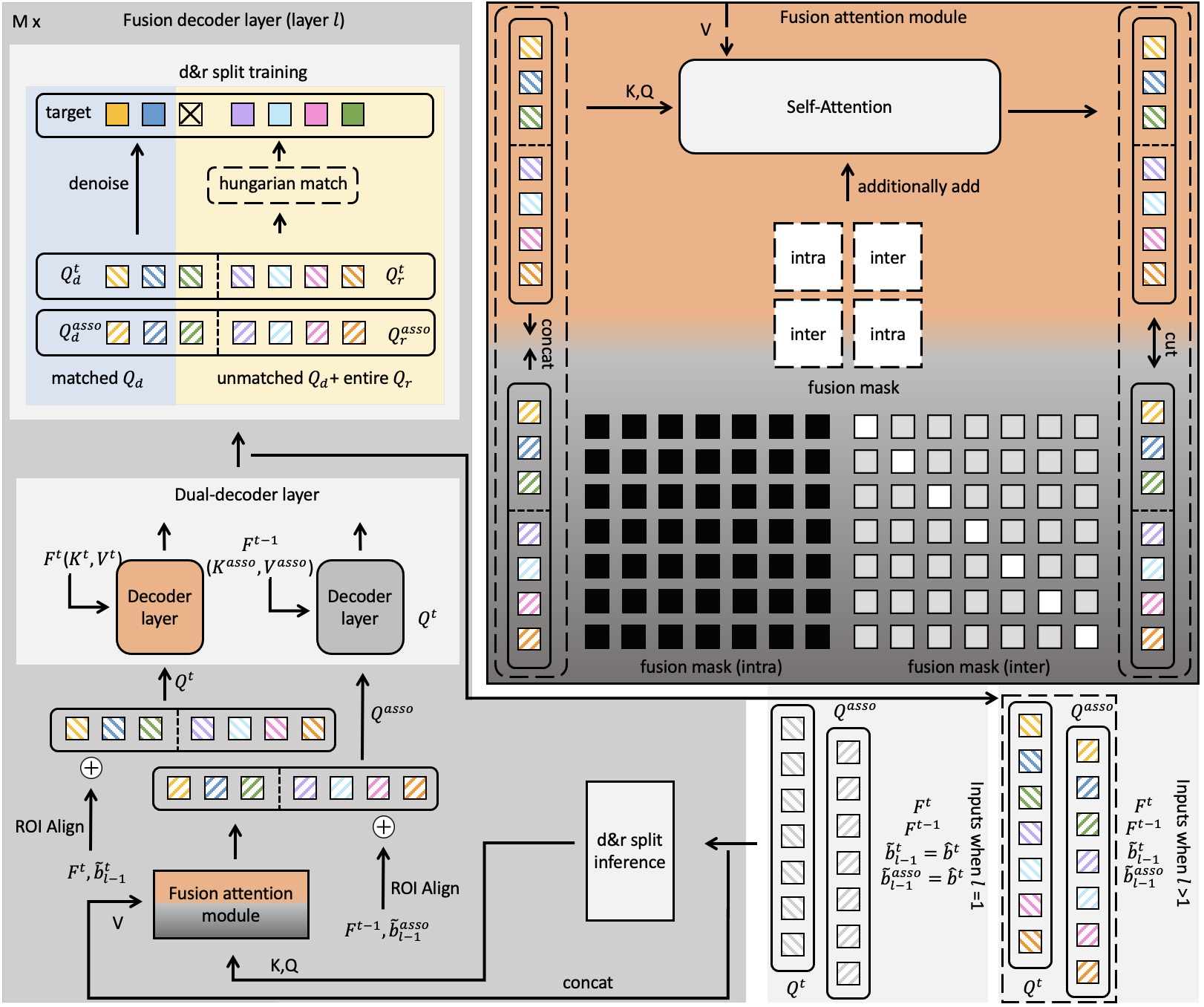}
	\end{center}
	\vspace{-2.5mm}
	\caption{\textbf{Roadmap of training fusion decoder layers.} The fusion decoder layer starts by splitting queries into denoising and rematching groups (indicated by a dotted line inside subsequent queries). Fusion between query pairs is performed afterward. Detailedly, the fusion mask completely blocks intraframe exchange while selectively allowing communication between frames (white means no mask, gray for partial mask). We ROIAlign encoded features guided by $ \tilde{b}_{l-1} $ of layer $l-1$ (or $\hat{b}^{t}$) before decoding. The aligned features are added to query pairs serving as extra semantic information. Additionally, we train categorized query pairs in a targeted manner.}
	\label{fig:decoder_transfiner}
	\vspace{-3.0mm}
\end{figure}

%According to figure \ref{fig:decoder_transfiner}, the structure of fusion decoder layer contains a fusion attention module and dual-decoder layer. %The former is designed to allow the detection and association clues to flow interchangeably, and the latter is used to focus on $Q^t$ and $Q^{asso}$, respectively.

\noindent \textbf{Dual-decoder layer.} TransFiner provides association clues $ \tilde{a}^t $ in the form of motions, i.e., offsets (if ideal) pointing from $ b^{t} $ to $ b^{t-1} $ ($ b $ for ground truth boxes). The \textit{iterative bounding box refinement} \cite{zhu2020deformable} mode of decoder works by iteratively correcting box predictions from the former decoder layer (through rectifications $ \Delta\tilde{b}^{t} $ from $Q^{t}$). We, inside query pairs, extend the mode by simultaneously rectifying (predicting) the attached bounding boxes in frame $ t-1 $ (through motions $ \tilde{a}^{t} $ from $Q^{asso}$), that is

\begin{subequations}
	\begin{align}
		&\tilde{b}^{t}_l = \Delta{\tilde{b}^{t}_l } + (\tilde{b}^{t}_{l-1} \ {\rm if}\  l>1 \ {\rm else} \ \hat{b}^{t} \label{eq:det box}) \\
		&\tilde{b}^{asso}_l = {\tilde{a}^{t}_l } + (\tilde{b}^{t}_{l-1}  \ {\rm if}\ l>1 \ { \rm else} \ \hat{b}^{t} \label{eq:asso box})
	\end{align}
\end{subequations}
where $ l $ ($1 \leq l \leq M$) is the layer index of fusion decoder. In general, $ \tilde{b}^{t} $ and $ \tilde{b}^{asso} $, connected by the motions $ \tilde{a}^{t} $ across frames, separately propagate through the dual-decoder structure of fusion decoder.

% TODO: improve the illustration of this para
\noindent \textbf{Fusion attention} is the self-attention mechanism with an additionally added fusion mask $ Mask_{fusion} $. Depicted as the block with a color ramp from orange to gray in Figure \ref{fig:decoder_transfiner}, fusion begins by concatenating the embeddings from query pairs (i.e., $ Q^t, Q^{asso} \in \mathbb{R}^{N \times d}  \stackrel{Concat}{\longrightarrow} Q^{fusion} \in \mathbb{R}^{2N \times d} $, $ d $ is the feature dimension). Self-attention is then performed on $ Q^{fusion} $ constrained by $ Mask_{fusion} \in \mathbb{R}^{2N \times 2N}$ to focus on exchange of cross-frame information. Thus 
$ Mask_{fusion} = \left[ m_{i,j} \right]_{2N \times 2N} $ satisfies
\begin{small}
	\begin{equation}
		\label{eq:fusionmask}
		m_{i,j} = \left\{
		\begin{aligned}
			0, 				 \    & {\rm if} \  (i - \dfrac{2N-1)}{2}) \times (j - \dfrac{2N-1)}{2} ) < 0 \ {\rm and} \  i {\rm \%} N = j {\rm \%} N; \\
			-\infty,      \    & {\rm if} \  (i - \dfrac{2N-1)}{2}) \times (j - \dfrac{2N-1)}{2} ) > 0; \\
			\beta,		   \    & {\rm otherwise}.
		\end{aligned}
		\right.
	\end{equation}
\end{small}
$ \beta $ is a hyperparameter introduced in the following.

Detailedly, sub masks of $ Mask_{fusion} $ can be categorized into two groups, namely $ Mask_{intra} \in \mathbb{R}^{N \times N} $ serving as the mask of intra-frame (top-left and bottom-right of $ Mask_{fusion} $), along with $ Mask_{inter} $, similarly. $ \left(\tilde{b}^{t}_l,\tilde{b}^{asso}_l\right) $, following Equation \ref{eq:det box} and \ref{eq:asso box}, are moved from $\tilde{b}^{t}_{l-1}$ through query pairs, which require each pair to pinpoint a specific object. It is for this reason that elements along the main diagonal of $ Mask_{inter} $ are emphasized more than others, offering larger room for each pair to determine its target ($ \beta $ in Equation \ref{eq:fusionmask} shows this attention difference, $ -10 $ is our default setting, more to refer to the discussion on Table \ref{tab:ablation}). In a nutshell, Fusion mask is designed to improve the match between query pairs while reserving space for retrieving \textit{extra} information.

% Unlike the tedious image feature extraction process of the transformer's encoder, the decoder embraces way various potentials, mainly for the existence of object query and its derivations (e.g., reference locations \cite{zhu2020deformable}, weighted mask \cite{gao2021fast,zhang2022accelerating}). Moreover, 
\subsection{Decoder initialization} 
\label{sec:method:initialization}
Query pairs of the fusion decoder greatly contribute to the object predictions. Hence, it is straightforward to consider integrating the original predictions $ \hat{o}^{t} $ into their initialization.

\noindent \textbf{Reference locations.} TransFiner fills initial reference locations of two recent frames $ {init\_ref}^t $ and $ {init\_ref}^{asso} $ with $ \hat{b}^{t} $ ($ init\_ref $ is the same as $ \tilde{b}_{l=0} $).

\noindent \textbf{Query pairs.} Some \cite{yao2021efficient,zhang2022accelerating} inject the query embeddings with encoded features from the regions of interest. As shown in Figure \ref{fig:decoder_transfiner}, we similarly ROIAlign \cite{he2017mask} the encoded features within reference locations $ \tilde{b}_{l-1} $ under layer $l$, resulting in $ 2N $ aligned feature maps. Afterward, extracting and combining the features from the sampling points of each feature map yields $ 2N $ distinct feature embeddings, which are then added to the corresponding query pairs.

\subsection{Query denoising \& query rematching} 
\label{sec:method:objectquerycategorizing}
Prediction is often categorized as good or bad based on its accordance with the supposed ground truth. The former usually takes less effort than the latter under refinement. In other words, a query initialized from the former usually has a closely related target, which may suffer from the instability of the Hungarian matching (i.e., target shift as the disturbance introduced in refinement, a similar question discussed in \cite{li2022dn}). Hence, we introduce \textit{denoising and rematching split} (d\&r split for short), including inference and training steps shown in Figure \ref{fig:decoder_transfiner}.

\noindent \textbf{Inference.} We distinguish a query for denoising or rematching by comparing its objectness score $ \hat{c}^t _i $ from accordingly initialized original prediction $\hat{o}^{t}_{i}$ with $ {thr}_{out} $ (e.g., 0.4). Afterward, we label queries by assigning the denoising embedding $ q_{d} $ to those associated with decent predictions, i.e., $ \left({Q}^t_d, Q^{asso}_d\right) $, and the rematching embedding $ q_{r} $ to those related to poor predictions, i.e., $ \left({Q}^t_r, {Q}^{asso}_r\right) $. There is a reminder that decoder performs identification over denoising and rematching at the first layer.

\noindent \textbf{Training.} After conducting the \textit{inference} step amid training, we further pre-determine the matched target-prediction pairs among $ Q_d $ following
\begin{small}
	\begin{equation}
		\label{eq:QDQR}
		{\rm Target}\left[\left({Q_d^{t}}\right)_m\right]= \left\{
		\begin{aligned}
			{b}^{t}_{\sigma\left(m\right)}, \
			{\rm if} \ & {\rm iou}\left({b}^{t}_{\sigma\left(m\right)},\hat{b}^{t}_{m} \right)\ >{thr}_{match}; \\
			\emptyset,   \ \ \ \ \ \     &{\rm otherwise}.
		\end{aligned}
		\right.
	\end{equation}
\end{small}
$\sigma$ is the optimal assignment from Hungarian match between decent predictions and targets. ${thr}_{match}$ is the threshold filtering denoising queries whose initialized locations intolerably deviate from targets even with high objectness scores.

In the subsequent layer-by-layer refinement, Hungarian matching is performed outside the matched $ Q_d $, leaving unmatched $ Q_d $ and the entire $ Q_r $ to search for the best-associated targets in each layer.

\section{Experiments}
\label{sec:exp}

\subsection{Datasets \& evaluation metrics}
\noindent \textbf{MOT.} In multiple object tracking, MOT benchmarks are generally used to evaluate the performance of trackers. We conduct experiments on the MOT16 and MOT17 \cite{milan2016mot16}, both including 7 training sequences and 7 test sequences. The final results reported in Section \ref{sec:exp:officialresults} are obtained through training on the entire train set (additionally with the validation set of CrowdHuman \cite{shao2018crowdhuman}) and evaluating on the test set officially under the private detection protocol. For the ablation study, we, following Centertrack \cite{zhou2020tracking}, split the official train set into two halves. The first half is used for training, while the second is for validation.

\noindent \textbf{CrowdHuman} \cite{shao2018crowdhuman} is a detection dataset filled with collections of images of the crowd, containing 15000 training images and 4370 validation images, which is widely used as a pre-training dataset for the MOT trackers.

\noindent \textbf{Metrics.} We demonstrate our results using the popular MOT evaluation metrics set CLEAR \cite{bernardin2008evaluating}, including Multiple-Object Tracking Accuracy (MOTA), Identity Switch (IDS), False Positive (FP), and False Negative (FN). Additionally, we report the IDentification F1 score (IDF1) \cite{ristani2016performance} and the Higher Order Tracking Accuracy (HOTA) \cite{luiten2021hota}, which is the geometric mean of two sub-metrics comprising Association Accuracy score (AssA) and Detection Accuracy score (DetA).

\subsection{Implementation details}
\noindent \textbf{Model.} We pick CenterTrack \cite{zhou2020tracking} as the original tracker in our experiments. For TransFiner, the backbone network is ResNet-50 \cite{he2016deep}, coupled with the \textit{twin} structure from a six-layer encoder and decoder of Deformable DETR \cite{zhu2020deformable}. The number of query embeddings is set to $ 300 $. %In MOT datasets, the bounding boxes fully cover the targets, which means parts of the objects have their box centers outside the images, making it suboptimal to directly predict the objects' centers, widths, and heights. Hence, following the solution in \cite{zhou2020tracking}, we also formulate the box representation set $ {b}=\left({x}, {y}, {ad}_{tp}, {ad}_{lf}, {ad}_{bt}, {ad}_{rt}\right) $. The last four respectively show the non-negative distance from the center to the top, left, bottom, and right edge of the bounding box. This allows more precise estimations even when objects are heavily cropped.

%\begin{comment}
\noindent \textbf{Decoder initialization.} Formalized in Section \ref{sec:Preliminaries}, TransFiner's decoder outputs 
$ \left\{ \tilde{y}^{t}_{i} \right\}_{i=0}^{N-1} $, while its initialization input is 
$ \left\{ \hat{o}^{t}_{i} \right\}_{i=0}^{K-1} $. Obviously, the mismatch between $ K $ and $ N $ raises the question of how to perform a one-to-one assignment at the beginning of the object querys' initialization. Here we provide a feasible solution. Following the categorizing standard in Section \ref{sec:method:objectquerycategorizing}, we address this first by separating $ \hat{o}^{t} $ into set $ \left(\hat{o}^{t}_{d}, \hat{o}^{t}_{r}\right) $, and there are respectively $ K_{d} $ and $ K_{r} $ elements in $ \hat{o}^{t}_{d} $ and $ \hat{o}^{t}_{r} $. Next, we obtain the sequence by linking $ \hat{o}^{t}_{d} $ with $ \lceil \dfrac{N-K_{d}}{K_{r}} \rceil $ times repeated $ \hat{o}^{t}_{r} $. The sequence is then clipped to that of length $ N $.
%\end{comment}

% need to be further determined when experiments are done (ciou mention?)
\noindent \textbf{Training settings.} Images are resized to $ 672 \times 1184 $ as inputs. Following the coefficients of Hungarian loss in \cite{carion2020end}, which are 2, 5, and 2 for $ \lambda_{cls} $, $ \lambda_{L_1} $, and $ \lambda_{iou} $, respectively. We use the last two for the loss calculation on detection boxes $ \tilde{b}^{t} $ in Equation \ref{eq:det box}, while estimation of association boxes $ \tilde{b}^{asso} $ (from Equation \ref{eq:asso box}) are trained under the same coefficients divided by 5. Due to GPU memory limitation, the batch size is set to 8, with gradient accumulation amid every two iterations and simulating a 16-batch setup. Overall, we use 2 NVIDIA RTX 3090 GPUs with batch size 8, optimizer AdamW \cite{loshchilov2018decoupled}, and the initial learning rate $ 2e-4 $. TransFiner is first pre-trained on the CrowdHuman train set \cite{shao2018crowdhuman} for 95 epochs, with learning rate dropping to $ 2e-5 $ after 50 epochs. We then train the TransFiner on both MOT \cite{milan2016mot16} and CrowdHuman validation set \cite{shao2018crowdhuman} for another 130 epochs with learning rate decreasing by 10 at 100-th epoch. 

\begin{comment}
	\begin{table}[!htbp]
		\begin{center}
			%\resizebox{\textwidth}{!}{
				\begin{tabular}{cccccc}
					\toprule[1.5pt]
					\textbf{Method} 
					& MOTA$\uparrow$  &IDF1$\uparrow$ 
					& FP$\downarrow$ &FN$\downarrow$& IDS$\downarrow$\\
					\midrule
					
					CenterTrack \cite{zhou2020tracking}
					& 67.8  &64.7   & 18498 &160332 & 3039\\ %mot17
					
					CenterTrack+TransFiner 
					& 71.5  &66.8  & 29283 &128665 & 3056 \\ %mot17
					\bottomrule[1.5pt]
				\end{tabular}%}
			\caption{Test set results of CenterTrack \cite{zhou2020tracking} without and with TransFiner on MOT17.}
			\label{tab:MOTimprove}
		\end{center}
		% data needed to be further determined
	\end{table}
\end{comment}

\begin{table}[t!]
	\centering
	\footnotesize
	\setlength\tabcolsep{4.0pt}
	\resizebox{\textwidth}{!}{
		\begin{tabular}{l|cccccccc}
			\toprule[1.5pt]
			\textbf{Method}&\textbf{IDF1}  $\uparrow$ & \textbf{MOTA} $\uparrow$ & \textbf{HOTA} $\uparrow$ & \textbf{DetA} $\uparrow$& \textbf{AssA} $\uparrow$  & \textbf{IDsw} $\downarrow$ & \textbf{FP} $\downarrow$ & \textbf{FN} $\downarrow$  \\\hline
			\multicolumn{9}{c}{\textbf{MOT16} } \\\hline
			%FairMOT~\cite{zhang2020fairmot}  & \textbf{72.3} & 69.3 & \textbf{58.3} & 58.8 & \textbf{58.0}  & \textbf{815} & 13501 & 41653\\
			TubeTK~\cite{pang2020tubetk}   & 62.2 & 66.9 & 50.8 & 55.0 & 47.3 & 1236  & 11544 & 47502 \\
			Chain-Tracker~\cite{peng2020chained} & 57.2 & 67.6 & 48.8 & 55.0 & 43.7 & 1897 & \color{blue}8934 & 48350 \\ 
			TraDeS~\cite{wu2021track}  & 64.7 & 70.1 &53.2 &56.2 & 50.9 & 1144 & \color{red}{8091} & 45210 \\
			QuasiDense\cite{pang2021quasi} &  67.1 & 69.8 & 54.5 &56.6 & \color{blue}52.8 &  1097  & 9861 & 44050 \\
			MeMOT~\cite{cai2022memot} & \color{red}{69.7} & 72.6 & \color{red}{57.4} & - & \color{red}{55.7} & \color{red}{845} & 14595 & \color{red}{34595} \\
			PatchTrack~\cite{chen2022patchtrack} & 65.8 & \color{red}{73.3} &54.2 & \color{red}{59.6} & 49.7 & 1179 & 10660 & \color{blue}36824 \\
			CenterTrack\textbf{+TF} (ours) & \color{blue}67.6 & \color{blue}73.0 & \color{blue}55.1 & \color{blue}58.6 & 52.2 & \color{blue}976 & 10463 &37723 \\\hline

			\multicolumn{9}{c}{\textbf{MOT17}} \\\hline
			%FairMOT \cite{zhang2020fairmot}  & \textbf{72.3} & 73.7 & \textbf{59.3} &\textbf{60.9} & \textbf{58.0} & 3303  & 27507 & 117477\\
			TraDeS~\cite{wu2021track} &  63.9 & 69.1 & 52.7 &55.2 & 50.8 & 3555 & \color{blue}20892 & 150060 \\
			QuasiDense\cite{pang2021quasi} &  66.3 & 68.7 & 53.9 &55.6  & \color{blue}52.7&  3378  & 26589 & 146643 \\
			TransTrack~\cite{sun2020transtrack}  &  63.9 & \color{red}{74.5}  & 53.9 &\color{red}{60.5}  & 48.3 & 3663 & 28323 & \color{red}{112137} \\
			TransCenter~\cite{xu2021transcenter}  & 62.2 & 73.2 & \color{blue}54.5 &\color{blue}60.1 & 49.7 & 4614 & 23112 & 123738 \\
			TubeTK~\cite{pang2020tubetk} & 58.6 & 63.0 & 48.0 & 51.4 & 45.1 & 4137  & 27060 & 177483 \\
			Chain-Tracker~\cite{peng2020chained} & 57.4 & 66.6 & 49.0 &53.6 & 45.2 & 5529 & 22284 & 160491 \\
			TrackFormer~\cite{meinhardt2021trackformer}  & 63.9 & 65.0 & - &- & -& 3258 & 70443 & 123552 \\
			MeMOT~\cite{cai2022memot}  & \color{red}{69.0} & 72.5 &\color{red}{56.9} & -& \color{red}{55.2} & \color{red}{2724} & 37221 & \color{blue}115248 \\ 
			PatchTrack~\cite{chen2022patchtrack} & 65.2 & \color{blue}73.6 & 53.9 &59.4 & 49.3 & 3795 & 23976 & 121230 \\
			CenterTrack \cite{zhou2020tracking} & 64.7 & 67.8 & 52.2 &53.8 & 51.0 & \color{blue}3039  & \color{red}{18498} & 160332\\
			CenterTrack\textbf{+TF} (ours) & \color{blue}66.8 & 71.5 & \color{blue}54.5 &57.5 & 52.0 & 3056 & 29283 &128665\\
			\hline
			\bottomrule[1.5pt]
	\end{tabular}}
	\vspace{-2.0mm}
	\caption{\textbf{Evaluation results on MOT challenge datasets (private detection).} The \textbf{TF} stands for TransFiner. The best result in each column is marked in red and in blue for the second-to-best.}
	\label{tab:MOTexp}
	\vspace{-4.0mm}
	% data needed to be further determined
\end{table}

\begin{figure}[t!]
	\begin{center}
		\includegraphics[width=0.9\linewidth]{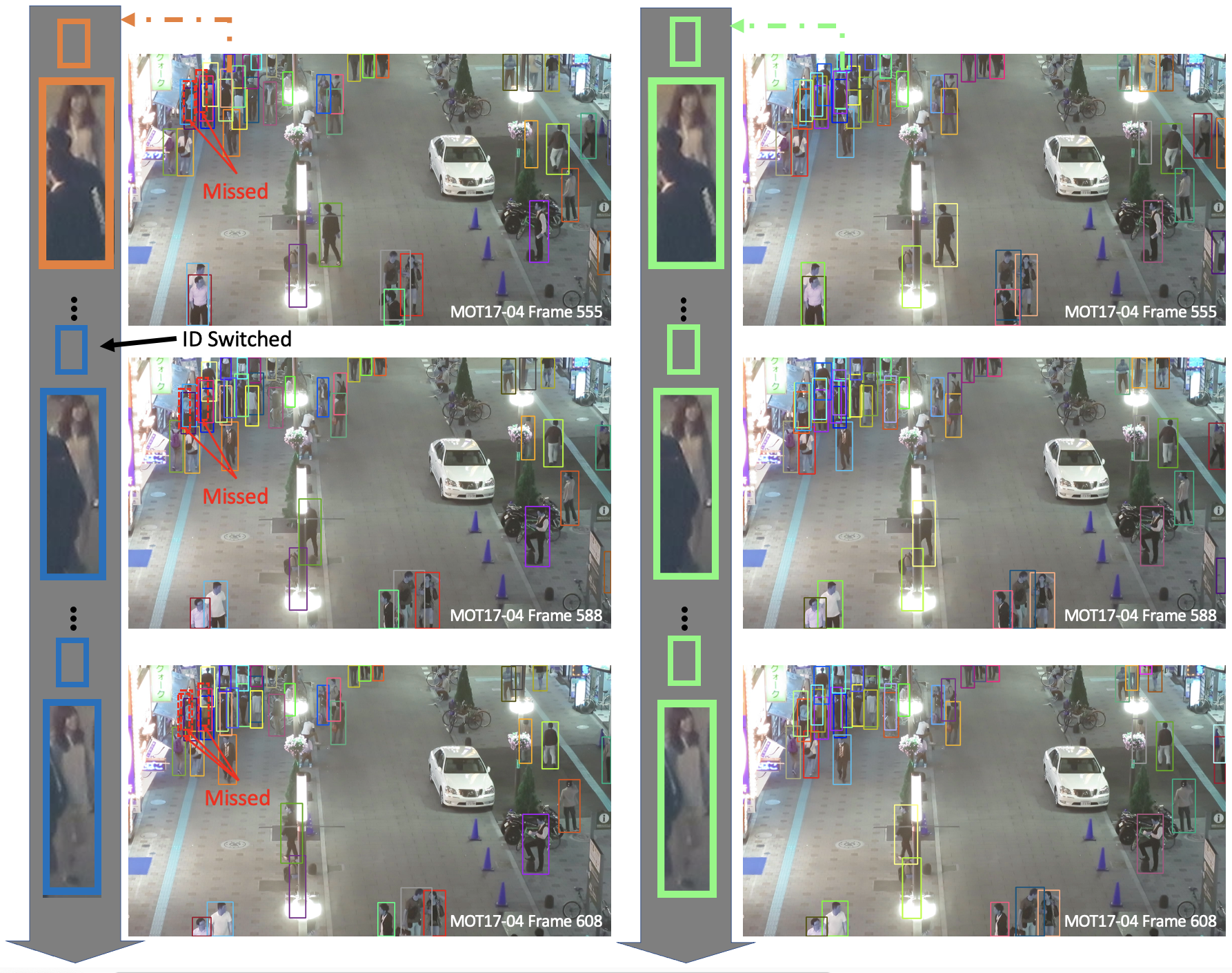}
	\end{center}
	\vspace{-1.0mm}
	\caption{\textbf{Case study.} Examples of CenterTrack (left column) refined by TransFiner (right column). Tracks are marked by color. The big black arrow depicts a tracklet of identical objects across frames. Under CenterTrack, the pedestrian in the orange box at Frame 555 now appears in a blue box since Frame 588. TransFiner, on the other hand, handles the identity switch originally introduced by this target via continuous tracks with green boxes. Moreover, additional annotations in red denote objects that CenterTrack ignores, while TransFiner fixes them.}
	\vspace{-2.5mm}
	\label{fig:visualization}
\end{figure}
\subsection{Benchmark results}
\label{sec:exp:officialresults}

\noindent \textbf{Improvement under TransFiner.} CenterTrack officially reports results on the MOT17 benchmark, where we have a detailed look. As shown in Table \ref{tab:MOTexp}, refinement by TransFiner shows a comprehensive improvement (+2.1\% IDF1 and +3.7\% MOTA). This benefits from distinct focuses of query pairs over targets, contributing to apparent refinements on FN (decreasing by 31667), while IDsw virtually stays intact (from 3039 to 3056). An example depicted in Figure \ref{fig:visualization}.

\noindent \textbf{MOT16 \& MOT17.} Table \ref{tab:MOTexp} demonstrates results reported on MOT16 and MOT17 test datasets. In MOT16, we chiefly compare enhanced CenterTrack with two other transformer-based trackers, namely PatchTrack \cite{chen2022patchtrack} and MeMOT \cite{cai2022memot}, which respectively obtain state-of-the-art performance in detection and association. Improved CenterTrack achieves comparative detection performance (73.0\% MOTA and 58.6\% DetA), with 0.3\% less MOTA and 1.0\% fewer DetA than PatchTracks. Alternatively, we better associate objects than PatchTrack, relying on the informative motions from query pairs, but still underperform MeMOT on IDF1 (67.6 vs. 69.7) and AssA (52.2 vs. 55.7), possibly due to our local linkage (performing on two continuous frames). In MOT17, CenterTrack powered by TransFiner embraces second-to-best tracking ability, surpassing most transformer-based approaches like TransTrack \cite{sun2020transtrack}, TransCenter \cite{xu2021transcenter}, Trackformer \cite{meinhardt2021trackformer} and PatchTrack \cite{chen2022patchtrack}. In addition, CenterTrack with TransFiner detects well (57.5\% DetA) but is inferior to several SOTA transformer-based trackers. It is probably because query pairs restrict the prediction of objects on the current frame if they are out of scope on the previous frame.

\subsection{Ablation study}
\label{sec:ablation}
% TODO: may be can add track query without output objective just pass through the fusion for "memory" purpose.
% filling this with after finish coding in TODO above
%\noindent \textbf{Only object queries.}
We test our design choices with the same model combination (CenterTrack and TransFiner) in Section \ref{sec:exp:officialresults} on the train-val split of the MOT17 train dataset.

\begin{table}[t]
	\begin{center}
		\resizebox{8cm}{!}{
			\begin{tabular}{cccccc}
				\toprule[1.5pt]
				Ablation &Choice & MOTA  &IDF1 & HOTA &AssA\\ \hline
				
				\rowcolor{LightCyan}&Single & 62.3  &59.0  & 48.6 &44.8 \\
				\rowcolor{LightCyan}\multirow{-2}*{Decoder structure} &*Fusion & \textbf{70.1} &\textbf{74.0}  & \textbf{60.6} &\textbf{63.0} \\ 
				
				\midrule
				\rowcolor{black!12}& w/ $\star$back refer & 69.8  &71.5  & 59.2 &60.0 \\ 
				\rowcolor{black!12}&w/o d\&r split &69.0   &72.6  &59.8  &61.6 \\ 
				\rowcolor{black!12}&w/o d\&r embeddings &68.9   &72.8  & 59.3 &60.5 \\ 
				\rowcolor{black!12}\multirow{-4}*{Refinement tactic} &*Vanilla & \textbf{70.1}  &\textbf{74.0}  & \textbf{60.6} &\textbf{63.0} \\
				
				\midrule
				\rowcolor{green!12}&0 &69.5  &71.8  & 58.8 & 59.5 \\
				\rowcolor{green!12}&-5 & \textbf{70.5}  &73.0  & 60.0 &61.1 \\ 
				\rowcolor{green!12}&*-10 & 70.1  &\textbf{74.0} & \textbf{60.6} &\textbf{63.0} \\ 
				\rowcolor{green!12}\multirow{-4}*{Hyperparameter $\beta$} &-$\infty$ & 69.5  &73.7  & 60.1 &62.0 \\ 
				
				\midrule
				\rowcolor{yellow!12}&*Center+Box & \textbf{70.1} &\textbf{74.0}  & \textbf{60.6} &\textbf{63.0} \\
				\rowcolor{yellow!12}&Center &69.0   &67.5 &56.6 &55.4 \\
				\rowcolor{yellow!12}\multirow{-3}*{Motion}&\XSolidBrush & 67.9  &65.7  &55.5  & 53.7 \\ 
				
				\hline 
				-&Baseline & 66.2  &69.4  & - &- \\
				
				\bottomrule[1.5pt]
		\end{tabular}}
		\caption{\textbf{Ablation studies on the MOT17 validation set.} * means our default settings.  An experimental attempt \textit{back refer} in Section \ref{sec:ablation} is indicated by $\star$. \textbf{Baseline} is the tracking performance of CenterTrack \cite{zhou2020tracking} under the same experiment settings. We explore design options on \textbf{decoder structure} (single-decoder structure fails), \textbf{refinement tactic} (d\&r split boosts refinement, and back refer drags it down), fusion mask \textbf{hyperparameter $\beta$} ($\beta=-10$ balances detection and association), and \textbf{motion} (box motion is critical in association). Color blocks with the best results are bolded.}
		\vspace{-2.0mm}
		\label{tab:ablation}
	\end{center}
	\vspace{-2.5mm}
\end{table}
% decoder focuses on the box detection without motion branch
\noindent \textbf{Decoder structure.} Fusion attention module and dual-decoder are layered repeatedly to form the \textit{fusion decoder}. Additionally, we receive the \textit{single} version by throwing fusion attention and the decoder focusing on $Q^{asso}$. Straightforwardly, refining with TransFiner built on \textit{single} merely redetects the objects of the current frame with specific decoder initialization. The results shown in the blue block of Table \ref{tab:ablation} suggest that the information fusion, as well as motion estimations, play a crucial role in MOT refinement. We observe fusion decoder elevates association significantly (15.0\% improvements on IDF1 and $ \sim $20.0\% increases on AssA compared with single decoders), indicating motions from query pairs of fusion decoder are robust in linking objects across frames.

% initialization for the decoder % reconstruct&rematch training strategy removal; 
\noindent \textbf{Refinement tactic.} We begin by exploring the initialization with back referring. Next, we discuss the ablations on the d\&r split of queries.

To further leverage $ \hat{o}^{t} $ during initialization of decoder, we attempt to extend the locations assignment in Section \ref{sec:method:initialization} by back referring $ {init\_ref}^{asso} $ through $ \hat{a}^t  $ instead of putting $ {init\_ref}^{asso} $ identical to $ {init\_ref}^{t} $. Specifically, back referring derives the reference locations of the previous frame through $ \hat{b}^{t}, \hat{a}^t $. Here we consider $ \hat{a}^t $ as backward motions. In this case, back referring is achieved via $ {init\_ref}^{asso} = {init\_ref}^t + \hat{a}^{t} $. The effectiveness of \textit{back refer} can be seen in the gray block of Table \ref{tab:ablation}, which shows overall performance degradation. We conclude two reasons for this: (1) Motions $\hat{a}^{t}$ from objects whose objectness scores are ucertain usually have a significant bias, deteriorating refinement by acting as \textit{unhealthy noises}; (2) Query pairs and the fusion mask allow for gradual adjustment of position pairs, discouraging excessive locations assignment beforehand.

For ablation studies on the d\&r split, we drop it from the vanilla. The 2nd row of the gray block in Table \ref{tab:ablation} shows that this lowers the model performance for, probably, pushing TransFiner to treat original predictions equally, without special attention to tough ones. In addition, we trial d\&r split lacking embeddings labeling denoising and rematching queries (i.e., without $ q_{d} $ and $ q_{r} $). This, however, further degrades TransFiner. Part of the reason is that little information hints at the queries with different refinement purposes when functioning.
\begin{comment}
	\begin{itemize}
		\item \textit{Motion.} In this case, \textit{back refer} is achieved via 
		\begin{equation}
			{init\_ref}^{t-1}_i = {init\_ref}^t_i + \hat{a}^{t}_i
		\end{equation}
		
		\item \textit{Previous Frame Bounding Box Predictions.} Matching the $ {init\_ref}^t $ and $ \hat{a}^{t} $ through hungarian match returns the matched pairs  $ \left\{  \left({init\_ref}^t_{i}, \hat{a}^{t}_{\sigma\left(i\right)} \right)\right\}_{i=0}^{N-1} $. \textit{Back refer} then conducts
		\begin{small}
			\begin{equation}
				{init\_ref}^{t-1}_i= \left\{
				\begin{aligned}
					\hat{a}^{t}_{\sigma\left(i\right)}, \
					{\rm if} \ & {\rm iou}\left(\hat{a}^{t}_{\sigma\left(i\right)},{init\_ref}^t_i \right)\ >{thr}_{similar}; \\
					{init\_ref}^t_i ,		   \     &{\rm otherwise}.
				\end{aligned}
				\right.
			\end{equation}
		\end{small}
		
		where $ {thr}_{similar} $ is the similarity threshold between the current prediction and the previous prediction assigned to $ {init\_ref}^{t-1}_i $.
	\end{itemize}
\end{comment}

% value for the mask (10, -inf, 0)
% 1. direct without fusion layer replaced with residual connection for interchangeable pre and current; 2. with the emphasis on attached queries while suppressing the theoretically unrelated queries
\noindent \textbf{Hyperparameter $\beta$.} Green rows of table \ref{tab:ablation} show optimization performances under various choices of $\beta$. $\beta=0$ leads to an obvious decline in association (reducing IDF1 by 2.2\% and 3.5\% for AssA from the default setting). In contrast, detection and association suffer slightly when $\beta=-\infty$, dropping from the vanilla by 0.6\% MOTA and 0.3\% IDF1. Moreover, we observe mild overall improvement when placing $\beta$ to a moderate value (e.g., $-10$). An intuitive illustration is that a suitable value of $\beta$ properly weighs interactions between queries outside and inside their \textit{in-couples}, where queries are dynamically and controllably fitted.

% output without motion
\noindent \textbf{Motion.} Transfiner evaluates motions in the form of centers and boxes of objects from the present to the last frame. According to the yellow chunk of Table \ref{tab:ablation}, we observe a considerable gap with and without box motions in the association (74.0\% IDF1 vs. 67.5\% IDF1 and 63.0\% AssA vs. 55.4\% AssA), considering box motions are more distinctive in crowded scenarios.

\subsection{Limitations} 
TransFiner performs on local tracking (within adjacent frames), limiting refinement when the targets are under long-term occlusions. To address these, the design of a prediction error buffer (e.g., contains the TransFiner's predictions crossing the border of d\&r split), along with a stronger query interaction mechanism, may help improve this defect. In addition, although TransFiner leverages initial tracking non-trivially, how to better \textit{semantically} joint inputs (e.g., frames) and outputs (i.e., original predictions) space is an open question.

\section{Conclusion}
We present TransFiner, a generic post-refinement framework for MOT. By adapting transformer for refinement, we, using the original tracker, simply consider the predicted locations and objectness scores. TransFiner fully exploits initial predictions, locations guide the extraction of image features for query pairs and scores are used to group pairs for targeted rectification. Labeled query pairs, highly representing original predictions, deeply combine the input and output space for refinement via propagating through fusion decoder. Our tracker-booster achieves impressive refinement outcomes on MOT16 and MOT17 benchmarks.

\bibliographystyle{splncs04}
\bibliography{main}
\end{document}